\RequirePackage{fix-cm}
\documentclass[twocolumn]{svjour3}         
\smartqed  
\usepackage{graphicx}
\usepackage{amsmath}
\usepackage{multirow}
\usepackage{pifont}
\usepackage{graphicx}
\usepackage{amsmath}
\usepackage{amssymb}
\usepackage{array}
\usepackage{multirow, nicefrac}
\usepackage{pifont}
\usepackage{cite}
\usepackage{bbm}
\usepackage{arydshln}
\usepackage{xcolor}
\usepackage{tablefootnote}
\usepackage{threeparttable}
\definecolor{green}{rgb}{1,0,0}

\def\T{{\!\top}}
\def\bW{{\mathbf W}}
\def\bM{{\mathbf M}}
\def\bx{{\mathbf x}}
\def\ba{{\mathbf a}}
\def\bb{{\mathbf b}}

\def\eg{\emph{e.g.}}

\def\etal{\emph{et al.}}

\def\etal{{\em et al.\/}\, }

\def\0{{\bf 0}}
\def\1{{\bf 1}}

\def\bM{{\bf M}}

\def\bW{{\bf W}}

\def\ba{{\bf a}}
\def\bb{{\bf b}}

\def\bx{{\bf x}}

\def\st{\mbox{s.t. }}

\usepackage{listings,xcolor} 
\usepackage{tcolorbox}

\newcommand{\lstfont}[1]{\color{#1}\ttfamily}
\newcommand{\code}[1]{{\ensuremath{\tt #1}}} 

\usepackage{xspace}
\usepackage[colorlinks,linkcolor=red,anchorcolor=blue,citecolor=purple,CJKbookmarks=True]{hyperref}

\usepackage{marvosym}

\begin{document}
\begin{sloppypar}
\title{A Dynamic Feature Interaction Framework for Multi-task Visual Perception 
\thanks{YX, HC contributed equally. 
Part of this work was done when YX was visiting Zhejiang University. 
}
}

\author{Yuling Xi,
        Hao Chen,
        Ning Wang,
        Peng Wang,
        Yanning Zhang,
        Chunhua Shen,
        Yifan Liu
}
\authorrunning{Y. Xi, H. Chen, N. Wang, P. Wang, Y. Zhang,    
 C. Shen, Y. Liu}

\institute{
Yuling Xi, Ning Wang, Peng Wang, Yanning Zhang({\color{blue}{\Letter}})
              \at
              School of Computer Science; and Ningbo Institute, Northwestern Polytechnical University, China \\
              \email{ {\tt \{xiyuling, ningw\}@mail.nwpu.edu.cn}, \\ 
              {\tt \{peng.wang, ynzhang\}@nwpu.edu.cn}} 
           \and
           Hao Chen, Chunhua Shen \at
           Zhejiang University, China \\
           \email{{\tt stanzju@gmail.com}, {\tt chunhuashen@zju.edu.cn} } 
           \and
           Yifan Liu \at 
           The University of Adelaide, Australia              \\
              \email{\tt yifan.liu04@adelaide.edu.au} 
}

\date{\today}

\maketitle

\begin{abstract}
Multi-task visual perception has 
a wide range of 
applications in scene understanding such as  autonomous driving. In this work, 
we devise an efficient unified framework to 
solve 
multiple common perception tasks, 
including instance segmentation, semantic segmentation, monocular 3D detection, and depth estimation. 
Simply sharing the same visual feature representations for these tasks impairs the performance of tasks, while independent task-specific feature extractors 
lead to 
parameter redundancy and latency. 
Thus,  we design two feature-merge 
branches to learn feature basis, which can be useful to, and thus shared by, multiple perception tasks. 
Then, each task 
takes the corresponding feature basis as the input of the prediction task head to 
fulfill 
a specific task. In particular, 
one feature merge branch is designed for instance-level recognition the other for dense predictions. 
To enhance inter-branch communication, the instance branch 
passes pixel-wise spatial information of each instance to the dense branch using efficient dynamic convolution weighting. 
Moreover, a simple but effective dynamic routing mechanism  is proposed to isolate task-specific features and leverage common properties among tasks. Our proposed framework, termed D2BNet, demonstrates a unique approach to 
parameter-efficient predictions for 
multi-task perception. 
In addition, as tasks benefit from co-training with each other, our solution achieves \textit{on par} results on partially labeled settings on nuScenes and outperforms previous works for 3D detection and depth estimation on the Cityscapes dataset with full supervision.

\keywords{Multi-task perception \and dynamic routing \and 3D object detection \and panoptic segmentation \and  depth estimation}

\end{abstract}

\section{Introduction}

Modern computer vision applications often deal with multiple tasks simultaneously. For instance, an AR application may need joint semantic understanding and 3D scene reconstruction, and a self-driving car relies on object detection, road segmentation, and depth estimation. A unified compact multi-task model could significantly reduce computation time and model size, which is crucial for real-world applications.

Recent multi-task architectures either have task-specific branches with a shared 
encoder 
\cite{kendall2018multi,wang2022semi} that ignores the correlation 
of high-level features or rely on an additional shared branch~\cite{nie2018mutual,schon2021mgnet, Xu_2018_Padnet,mousavian2016joint} to leverage common representations, which leads to computational burden and parameter redundancy. 
As a result, such elaborately designed networks often aim at specific joint tasks, which can hardly transfer to new tasks. Therefore, a unified multi-task framework that can easily fit into different perception tasks and transfer to new tasks is necessary.

Perception tasks such as detection, segmentation, and depth estimation typically require both high-level context information and low-level fine-grained information to precisely describe details. As inherent 
synergy 
exists in most perception tasks, we resort to a unified two-branched network to extract basic features for different aims. In particular, an \textbf{instance branch} is responsible for learning multi-scale information to distinguish instance-level properties, \eg, semantic classes for mask and bounding boxes. A \textbf{dense branch} generates rich pixel-level representations for fine-grained localization, \eg, masks, box coordinators, and depth. Flexible combinations of these basic features ensure that our proposed unified two-branched network can handle multiple tasks jointly with minimal effort.

    Typical two-branched methods \cite{bolya2019yolact,chen2020blendmask,tian2020conditional} have made progress in instance segmentation by proposing a low-cost dynamic merging mechanism that aggregates instance-level information and high-resolution dense feature maps. This approach has been extended to panoptic segmentation~\cite{kirillov2019panoptic}. Wang \etal\ 
\cite{wang2021max} further enables interactions between these two branches with multiple transformer blocks. Instead of using computation-intensive self-attention blocks, we devise a lightweight module, 
termed 
Dynamic Message Passing (DMP), based on low-rank factorization to handle second-order information between two dense feature maps. It is parameter-efficient on high-dimensional feature maps and propagates spatial information across branches. This 
increases the performance of both dense and instance tasks with almost no additional computation cost. 

Previous work points out that conflicting loss functions may cause gradient updates in different directions for the shared parameters, making it difficult to properly optimize hard parameter sharing MTL. Similar observations also appear in our two-branched structure. If all the parameters are shared in the instance branch or the dense branch, performance decreases. Inspired by the ``cross-stitch'' unit \cite{misra2016cross}, which can automatically learn a combination of shared and task-specific representations, we propose a Dynamic Router (DR) that uses task and channel awareness to route task features and learn common representation implicitly. With limited extra computation, our model reaches competitive results on joint multi-task visual perception.

More specifically, our main contributions can be summarized as follows:
\begin{itemize}
    \item We propose a general and simple two-branched multi-task perception network, which breaks down tasks and groups same level features in a parameter-efficient way to maximize inter-task feature sharing. 
    To our knowledge, this is the first time that panoptic segmentation, monocular 3D detection, and depth estimation have been simultaneously addressed in a single network. 
    \item We propose Dynamic Message Passing (DMP) to communicate information across branches and tasks. Through a parameter- and com\-pu\-ta\-ti\-on-eff\-ici\-ent feature merging operation, our network interacts spatial information between branches. Moreover, given the sharing and conflicting relations among tasks, a task- and channel-aware Dynamic Router (DR) is proposed to isolate task-specific features and utilize common properties of the tasks. 
    \item We demonstrate significant improvements for multiple perception tasks under simple co-training strategies. Our framework achieves competitive results on nuScenes in a partially labeled setting and surpasses previous methods for 3D detection and depth estimation by a great margin on the 
    Cityscapes dataset in a fully labeled setting.
\end{itemize}

\section{Related Work}

\textbf{Multi-task visual perception} in scene understanding targets the problem of training relationships between tasks. 
Extensive multi-task perception works have resorted to a single branch network that usually focuses on low-level dense prediction tasks to investigate the relationship among fine-grained features~\cite{standley2020tasks,maninis2019attentive, xu2023demt}. Joint learning of instance-level and dense prediction tasks has also been studied. However, these approaches either use a unified hard parameter sharing approach and experience rapid performance degradation~\cite{kokkinos2017ubernet} or use a separate decoder for each task~\cite{teichmann2018multinet,kirillov2019panoptic,leang2020dynamic,yuan2022polyphonicformer}, which brings computational burden. 
To address the issues mentioned above, we first include \textbf{panoptic segmentation} as a typical joint instance and dense perception task. It tackles the problem of classifying every pixel in the scene by assigning different labels for different instances. Mainstream panoptic networks can be classified into two clusters, separate and unified approaches. Separate approaches rely on individual networks for stuff (semantic) and thing (instance) segmentation and focus on devising methods to fuse these two predictions~\cite{xiong2019upsnet,cheng2020panoptic} and resolving conflicts~\cite{yang2020sognet,qiao2021detectors}. Recent unified approaches includes PanopticFCN~\cite{li2021panopticfcn} and MaX-DeepLab~\cite{wang2021max}. both of which are box-free methods with end-to-end mask supervision. To remove box guidance, these methods rely on positional embedding and dynamic convolution on the entire feature map for each instance, both of which can be simplified with box-based methods. 

To study the relationship among multiple perception tasks and merge them into a unified framework, we also include \textbf{monocular 3D object detection} and \textbf{depth estimation} in our task set. Recovering 3D coordinates from a single image is known to be ill-posed and prone to overfitting. Normally monocular 3D object detection approaches directly regress 3D attributes with 2D detectors~\cite{wang2021fcos3d,liu2020smoke}, which is easily 
overfit
to object sizes~\cite{lian2021geometry}. Recent works on 3D object detection attempt to leverage depth information by adding a depth prediction layer to their 2D detector~\cite{park2021dd3d} to enhance the performance of monocular 3D detection. However, both these methods focus on the single detection task and only regress depth at the instance level, lacking fine-grained structure information and could hardly transfer to the depth estimation task.  
Therefore, we propose a multi-task visual perception framework to simultaneously address panoptic segmentation, monocular 3D object detection, and depth estimation tasks. Instead of relying on one branch to predict the result of each task, our framework naturally splits these tasks into two branches and groups same-level features of different tasks, which is parameter-efficient and easily deployed in real-world applications.

\textbf{Dynamic neural network} Nowadays, many networks have adopted some variants  of attention mechanism for both dense and instance prediction tasks. For dense prediction, it is used to learn a context encoding~\cite{zhang2018context} or pairwise relationship~\cite{zhao2018psanet}. Fully-convolutional instance prediction networks \cite{bolya2019yolact,chen2020blendmask,tian2020conditional} use a dynamic module to merge instance information with high-resolution features. This design usually involves applying a dynamically generated operator, which is essentially an inner product between two input features.

Different from previous dynamic modules that are only applied once during prediction, our approach aggregates multi-scale context information from the instance branch to refine the dense branch with an efficient dynamic operator. MaX-DeepLab~\cite{wang2021max} employs transformer modules~\cite{vaswani2017attention} for cross-branch communication, which is computation
-intensive thus the instance-level feature has to be sparse. Instead, we generate low-rank dynamic factors for the convolution layer. The formulation of the dynamic operator is closely related to linear 2nd-order operations such as Gated Linear Units~\cite{dauphin2017language} and Squeeze-and-Excite blocks~\cite{hu2018squeeze}. A key distinction is that our dynamic module is more similar to cross-attention than self-attention since it merges features from different branches. 
The dynamic routing mechanism masks out a subset of network connections, which has been used in various models for computation reduction~\cite{hou2020dynabert,li2020learning} and continual learning~\cite{wortsman2020supermasks}. Dynamically changing the weights of network operations can be regarded as a special case of feature-wise transformation~\cite{dumoulin2018feature-wise}. The most common form is channel-wise weight modulation in batch norm~\cite{li2018adaptive} and linear layers~\cite{perez2018film}. 
previous routing methods mainly concentrate on a single task, attempting to alleviate data variance or enhance the 
expression capability 
of the model in a single task. Our routing mechanism, however, is based on task correlations. Co-trained tasks implicitly share information, either at the instance-level or structural information.  Given a set of correlated tasks, we use a task- and channel-aware router to utilize the shard information and suppress conflicting features.

\section{Method}

\begin{figure*}[htbp]
\centering
\includegraphics[width=1.0\textwidth]{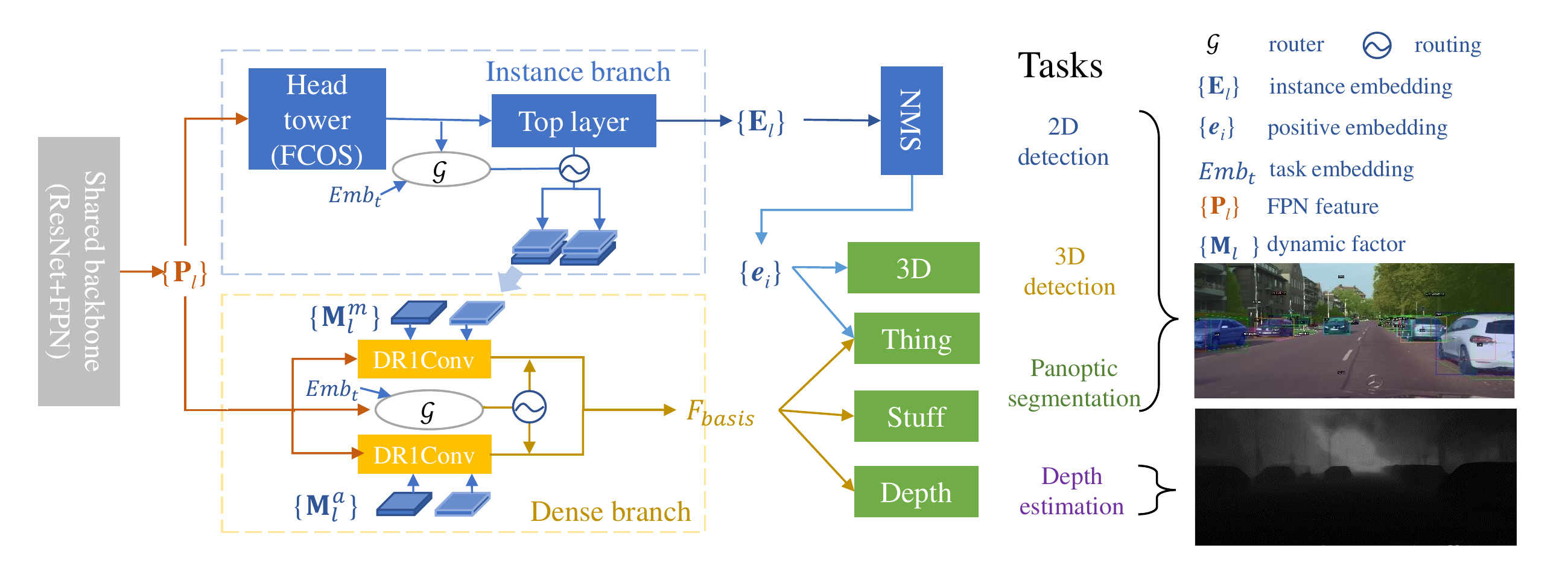}
\caption{\textbf{Overall pipeline.} 
Our model follows a typical two-branched framework with an instance branch extracting instance-level embeddings $\{ \mathbf E_l \}$ and $\{\mathbf M^{t}_l\}$, and a dense branch for predicting high-resolution dense feature map $\mathbf F$. The dense branch incorporates instance-level information with a dynamic operator DR1Conv. $\mathcal{G}$ is the proposed dynamic router. $Emb_t$ is task-aware information.
In the dense branch, common representations are leveraged to benefit each task by using the dynamic router $\mathcal{G}$. 
Compared to the hard sharing backbone with separable branches framework, D2BNet is more efficient and convenient for cross-task feature sharing.  
} 
\label{fig:main}
\end{figure*}

In this work, we resort to a two-branched network to unify instance and dense prediction tasks and maximize feature sharing, leveraging common representations to boost the performance of each task. In Sec.~\ref{sec:2b-arch}, we introduce the overall two-branched framework. To enhance information propagation between branches, we introduce an efficient dynamic module to pass on location information. To utilize common features in the same branch of different tasks, we devise a dynamic routing module to further selectively fuse channel-wise tasks features in Sec.~\ref{sec:3dy}. Task-specific prediction heads are described in Sec.~\ref{sec:head}. The overall pipeline is illustrated in Fig.~\ref{fig:main}.

\vspace{-10pt}

\subsection{Overall architecture}\label{sec:2b-arch}
Our framework comprises a feature extraction backbone, followed by two branches and separated task heads. 
The two-branched network includes an instance branch for higher-level contextual feature extraction and a dense branch for lower-level structural information prediction. 

The \textbf{instance branch} aims to generate instance-level semantic information and context using an arbitrary object detection decoder. We opt for a one-stage framework FCOS~\cite{tian2019fcos} due to its simplicity, and its multi-level architecture is convenient for investigating dynamic interaction between the dense branch and various instance-level feature maps.
For each instance $i$, our instance branch additionally generates an instance embedding $\boldsymbol{e}^{(i)}$ beyond FCOS. In this work, $\boldsymbol{e}^{(i)}$ contains all instance-level task embeddings, including 3D object attributes and things embeddings in panoptic segmentation, then joints with dense branch outputs to generate final predictions such as instance masks and 3D regression values. The $\boldsymbol{e}^{(i)}$ is generated by a top layer that is a single convolution layer added to the object regression tower to produce instance-wise contextual information. 
To propagate instance-level information between branches, this branch also generates a multi-scale conditional feature pyramid $ \{ \mathbf {M}_l \lvert l= 3; 4; \dots ; 7 \} $, which can be split along the channel for each task. 
Thus, given FPN output $\mathbf P_l$, the instance branch computes these features with the following equation:
\begin{align}
    \{\mathbf M_l, \mathbf E_l\} = \operatorname{Top}(\operatorname{Tower}(\mathbf P_l)),
    \;\;\;
    l=3, 4,\dots 7, 
\end{align}
where $\mathbf M_l$, $\mathbf E_l$ and $\mathbf P_l$ are tensors with the same spatial resolution. The densely predicted $\mathbf E_l$ along with other instance features such as class labels and bounding boxes are later filtered into a set containing only positive proposals $\boldsymbol{e}^{(i)}$. 
And $\mathbf M_l$ are further split into two dynamic tensors by channels for our dynamic operation called dynamic rank-1 convolutions (DR1Conv) to propagate location-aware information for dense prediction tasks. 
More details about DR1Conv will be described in Sec.~\ref{sec:loc}.

The \textbf{dense branch} preserves fine-grained image details to serve dense prediction tasks. Prior works on dense prediction tasks commonly use the largest resolution FPN feature map to generate per-pixel prediction, and ignore valuable semantic information of higher-level features. 
Our dense branch aggregates FPN features $\{\mathbf P_l\}$ with contextual features $\{\mathbf M_l\}$ into the final basis features $\mathbf F$ for dense prediction like an inverted pyramid. Starting from the highest level feature map with the smallest resolution, we use a dynamic operation to merge location-aware context from the instance branch. Channel- and task-aware information is also introduced to further interact with fine-grained features among dense prediction tasks. This merging operation is parameter-efficient and meanwhile enables our model to share tasks features to a great extent.

\subsection{Dynamic feature interaction modules}\label{sec:3dy}

To introduce communication within branches and tasks, we build dynamic modules upon our two-branched structure. Features to interact in the dynamic module include location-, channel- and task-aware information. 

\def\bW{{\mathbf W}}
\def\bM{{\mathbf M}}
\def\bx{{\mathbf x}}
\def\ba{{\mathbf a}}
\def\bb{{\mathbf b}}

\smallskip
\noindent \textbf{Dynamic Message Passing (DMP)}\label{sec:loc}
We design a dynamic interaction operation named DR1Conv for position-sensitive message passing between two branches. DR1Conv is inspired by BatchEnsemble~\cite{wen2020batchensemble}, which uses a low-rank factorization of convolution parameters for efficient model ensemble. 

\textbf{BatchEnsemble}~\cite{wen2020batchensemble} uses a low-rank factorization of convolution parameters for efficient model ensemble. 
One can 
factorize a weight matrix $\mathbf W'$ as a static matrix $\mathbf W$ and a 
low-rank matrix 
$\mathbf M$,

\begin{align}
    \mathbf W' = \mathbf W \odot \mathbf M, \mbox{  ~ ~ where } \mathbf M = \bb\ba^\T. \label{eq:rank1}
\end{align}

Here $\mathbf W', \mathbf W$, $\mathbf M\in\mathbb R^{m\times d}$, $\bb\in\mathbb R^m$, $\ba\in\mathbb R^d$ and $\odot$ is element-wise product. This factorization 
considerably reduces the number of parameters and 
requires less memory for computation. A forward pass with this dynamic layer can be formulated as

\begin{align}
{\mathbf y} &= {\mathbf W'}\mathbf x = (\mathbf W \odot \bb\ba^\T)
{\mathbf x} 
= (  \bW ( \bx    \odot    \ba    ) ) \odot \bb \label{eq:rankfwd}
\end{align}

where $\mathbf x\in\mathbb R^d$, $\mathbf y\in\mathbb R^m$ are the input and output vectors  respectively. Thus, this matrix-vector product can be computed as element-wise multiplying $\mathbf a$ and $\mathbf b$ before and after multiplying $\mathbf W$ respectively. This formulation also extends to other linear operations such as tensor product and convolution. Dusenberry \etal \cite{dusenberry2020efficient} use this factorization for efficient Bayesian posterior sampling in Rank-1 BNN.
Next, we extend this technique to convolutions to serve our purpose---to generate parameter-efficient \textit{dynamic convolution} modules for dense mask prediction tasks. 

We extend the factorization in Eq.~\eqref{eq:rank1} to convolutions. Different from BatchEnsemble and Rank-1 BNN
~\cite{wen2020batchensemble, dusenberry2020efficient}, our dynamic convolution is designed to be position-sensitive so that contextual information at different positions can be captured.  In other words, the rank-1 factors $\mathbf a$ and $\mathbf b$ 
have to preserve the location information of 2D images. 

In practice, we densely compute $\ba_{hw}$ and $\bb_{hw}$ for each location $(h, w)\in \mathbb [1,\dots,H]\times[1,\dots,W]$ as two feature maps $\mathbf A, \mathbf B \in \mathbb R^{C\times H\times W}$ whose spatial elements are the dynamic rank-1 factors. 

For simplicity, we first introduce the $1\times1$ convolution case. For each location $(h, w)$, we generate a different dynamic convolution kernel $\mathbf W'_{hw}\in\mathbb R^C$ 
from the corresponding locations of $\mathbf A$, $\mathbf B$. We apply dynamic matrix-vector multiplication at position $(h, w)$ as
\begin{align}
    \mathbf y_{hw} = \mathbf W'_{hw}\mathbf x_{hw} = (\mathbf W(\mathbf x_{hw}\odot \mathbf a_{hw}))\odot\mathbf b_{hw},
\end{align}
where $\mathbf a_{hw}, \mathbf b_{hw}\in\mathbb R^C$ 
are elements in the dynamic tensors $\mathbf A$ and $\mathbf B$. $\odot$ is element-wise multiplication. This can be interpreted as element-wise multiplying the context tensors before and after the static linear operator.

\begin{figure}[t!]
\centering
\includegraphics[width=0.7\linewidth]{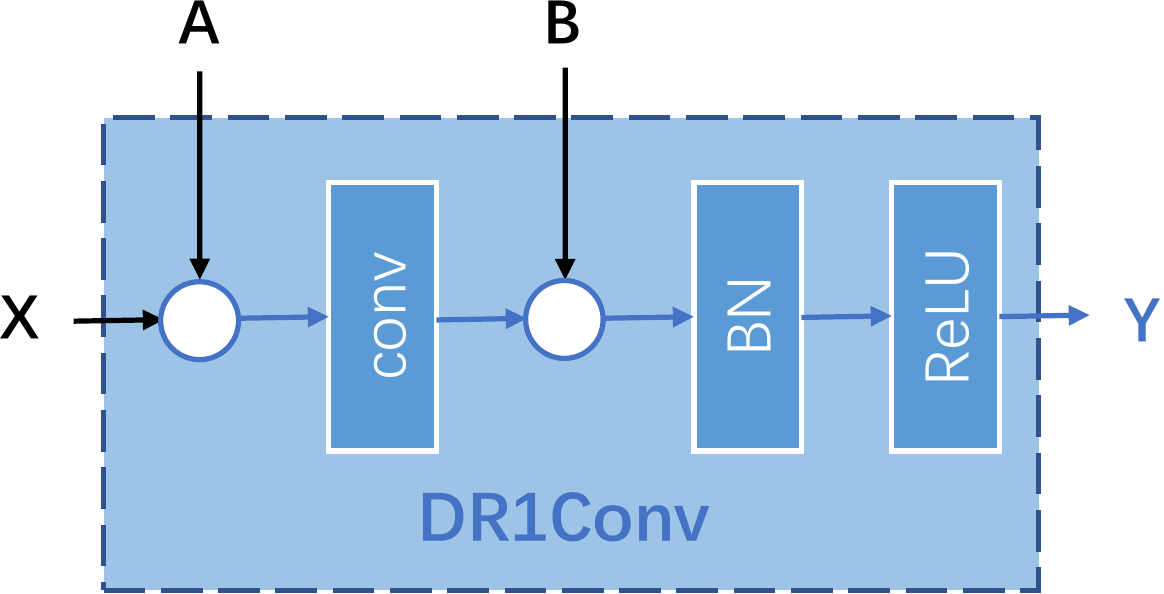}
\caption{Diagram of dynamic rank-1 convolution (DR1Conv). $\circ$ denotes element-wise multiplication. Tensors $\mathbf A$, $\mathbf B$ are the dynamic factors encoding the contextual information. Each modulated the channels of the feature before and after the convolution operation. $\mathbf X$ is the input and $\mathbf Y$ is the output. All tensors have the same size.}\label{fig:dr1conv}
\end{figure}

We then generalize this to arbitrary kernel shape $J\times K$. The dynamic rank-1 convolution (DR1Conv) $\operatorname{Conv}_{\mathbf W'}$ with static parameters $\mathbf W$ at location $(h, w)$ takes an input patch of $\mathbf X$ and dynamic features $\mathbf A$ and $\mathbf B$ and outputs feature $\mathbf y_{hw}$:
\begin{align}
\begin{split}
    \mathbf y_{hw} = \smash{\sum_j\sum_k}(\mathbf W[j, k]
    &(\mathbf X[h-j, w-k]\\
    &\odot \mathbf A[h-j, w-k]))\\
    &\odot\mathbf B[h-j, w-k].
\end{split}
\end{align}
We can
parallelize 
the element-wise multiplications between the tensors and compute DR1Conv results on the whole feature map 
efficiently. 
Specifically, we make dynamic convolution kernel position-sensitive by using two tensors ${\mathbf A,\mathbf B \in\mathbb R^{C\times H \times W}}$ generated from box regression tower with the same size as input feature $\mathbf X$. 
DR1Conv can be formulated as: 
\begin{equation}
    \mathbf Y = \operatorname{DR1Conv}_{\mathbf A, \mathbf B}(\mathbf X) = \operatorname{Conv}(\mathbf X\odot \mathbf A)\odot\mathbf B.
\label{eq:dr1conv}
\end{equation}
All tensors have the same size $C \times H \times W$. This is implemented as element-wise multiplying the dynamic factors $\mathbf A$, $\mathbf B$ before and after the static convolution respectively. 
The structure of DR1Conv is shown in Figure~\ref{fig:dr1conv}.

\begin{figure*}[t]
\centering
\includegraphics[width=0.615\textwidth]{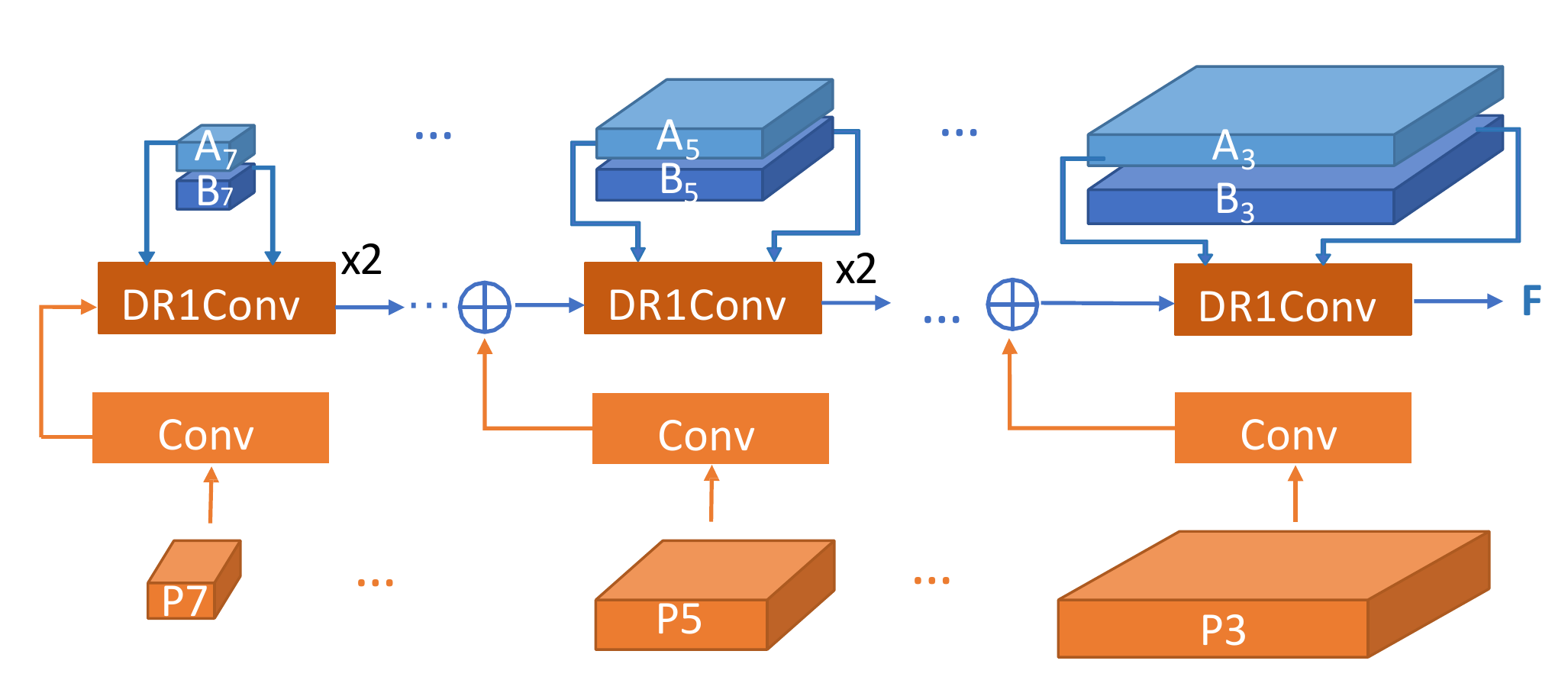}
\caption{The dense branch is an inverted pyramid network and our dynamic module is built upon it. $\{{\mathbf A_l,\mathbf B_l}\}$ contain position-sensitive information generated from multi-level features in the instance branch.  $\{\mathbf P_l\}$ are FPN features. }
\label{fig:dr1basis_1}
\end{figure*}

For the 
dense branch, we use DR1Conv to merge FPN outputs $\mathbf P_l$ and contextual features $\mathbf M_l$ from instance branch:
\begin{equation}
    \mathbf F_l = \operatorname{DR1Conv}_{\mathbf A_l, \mathbf B_l}(\operatorname{Conv}_{3\times3}(\mathbf P_l) + \operatorname{\uparrow_2}(\mathbf F_{l+1})),\label{eq:dr1basis1}
\end{equation}
where $\uparrow_2$ is upsampled by a factor of $2$. 
We first reduce channel width of $\mathbf P_l$ with a $3\times 3$ convolution, the channel width is kept the same throughout the computation. In practice, we found that for semantic segmentation, 64 channels are sufficient. The computation graph is shown in Fig.~\ref{fig:dr1basis_1}. 
After the last refinement, $ \mathbf F_3$ is output as the final $ \mathbf F$. This makes our dense branch very compact, using only $\nicefrac{1}{4}$ of the channels of the corresponding block compared with BlendMask~\cite{chen2020blendmask}. 

In our experiments, we found using DR1Conv makes our model $6\%$ faster while achieving even higher accuracy.
Note that we can parallelize the element-wise multiplications between the tensors and compute DR1Conv results on the whole feature map efficiently. In addition, it can be integrated into other fully-convolutional instance or panoptic segmentation networks. 
We argue that DR1Conv is essentially different from naive channel-wise modulation. The two related factors $\mathbf A$, $\mathbf B$ combine to gain much stronger expressive power while being very computationally efficient. 

\smallskip
\noindent \textbf{The Channel-wise Dynamic Routing (CDR) Mechanism 
} \label{sec:channel} Benefited from the aforementioned location-aware message passing within the two-branched architecture, we are able to investigate the  sharing of instance- and dense-level information across typical perception tasks. 
There is a common agreement that depth estimation and segmentation tasks both require structural information, and our experiments on co-training for panoptic segmentation and depth estimation revealed information sharing on both the instance and dense branches, with varying degrees of feature sharing across branches and feature scales.

Thus, to isolate task-specific features and leverage common representations, we introduce dynamic channel-wise routers to both the instance and dense branches at different scales for fine-grained feature interaction.
For each task in multi-task co-training, we assign it as the primary task and treat the others as secondary tasks. For instance, when segmentation and depth are co-trained, segmentation is the primary task and depth is the secondary task, and \textit{vice versa}. The objective of our 
routing mechanism 
is to separate task-specific features from shared features learned implicitly. We use different activation functions to differentiate routers for primary and secondary tasks. 
According to the phenomenon that features sharing varies on multiple scales, the channel 
routing 
is formulated as,
\begin{equation}
    \mathcal{G}_l= \{\sigma,\operatorname{softmax}\} (\operatorname{Conv}(\operatorname{GAP}(x_l)),\label{eq:gate}
\end{equation}
where $\sigma$ denotes the sigmoid function. $\operatorname{Conv}$ is a $1 \times 1$ convolution layer. The channel router leverages a sigmoid function to generate channel-wise activation to weigh the importance of the primary task feature per channel while using softmax to emphasize helpful channels and suppress harmful channels in the secondary task feature. 

The router outputs channel-wise attention scores to weight task features. In instance branch, routers take each level of box tower features as input $x_l$, routing task-specific contextual information $\{\mathbf M_l\}$ that we split into $\{\mathbf M^{m}_l, \mathbf M^{a}_l\}$ along channels. In the dense branch, we set two convolution layers in DR1Conv, denoting primary and secondary projections. Routers use the input features of DR1Conv in Eq.~\eqref{eq:dr1basis1} to generate task-specific weighting. 
The channel-aware routing result is formulated as,
\begin{equation}
    \mathcal{F}^{m}= \mathcal{G}^m \otimes F^{m} + \mathcal{G}^a \otimes F^{a}_,\label{eq:channel}
\end{equation}
where 
$\otimes$ is the channel-wise product. The output dimension is identical to the feature channel, routing features of primary and secondary tasks. Superscripts $m$ and $a$ denote primary and secondary tasks, respectively. $F^{\{m,a\}}$ are $\{\mathbf M^{m}_l, \mathbf M^{a}_l\}$ or DR1Conv projections features $\{\mathbf F^{m}_l, \mathbf F^{a}_l\}$ in each branch. For each task, we compute two routing scores that multiply to self and secondary features and combine them together to get $\{\mathbf M_l\}$ and $\{\mathbf F_l\}$ in each level. $\{\mathbf M_l\}$ and $\{\mathbf F_l\}$ are sent into Eq.~\eqref{eq:dr1basis1} to get final task-specific dense basis features. In this interactive mechanism, we maximize the extent of feature sharing across tasks and scales, and our model is able to utilize useful task features to enhance the expressivity of other tasks while using fewer computational resources and parameters.

\smallskip
\noindent \textbf{The Task-aware Dynamic Routing (TDR)
Mechanism 
} \label{sec:task} 
To bring more discrimination for different tasks, we introduce task-aware information to our dynamic router to further distinguish the task-specific representations that result in performance degradation. 

We add task ID embeddings to the Eq.~\eqref{eq:gate} to let features be more discriminative, thus the routing score is calculated as,
\begin{equation}
    \mathcal{G}= \{\sigma,\operatorname{softmax}\} (\operatorname{Conv}(\operatorname{GAP}(x)\oplus Emb_{\{t_m,t_a\}}),\label{task}
\end{equation}
where $\oplus$  is a concatenation operation, $Emb_{t}$ is the task embedding from a $1 \times 1$ convolution layer on the specific one-hot task id, and the parameter of this convolution is shared in all tasks. The shapes of GAP output and task id embedding are C ×1×1 and $\frac{C}{8}$ ×1×1. $\mathcal{F}^{m}_{3}$ is the final output $\mathbf F$ of each task in our multi-task network. 

\begin{figure*}[t]
\centering
\includegraphics[width=0.9\textwidth]{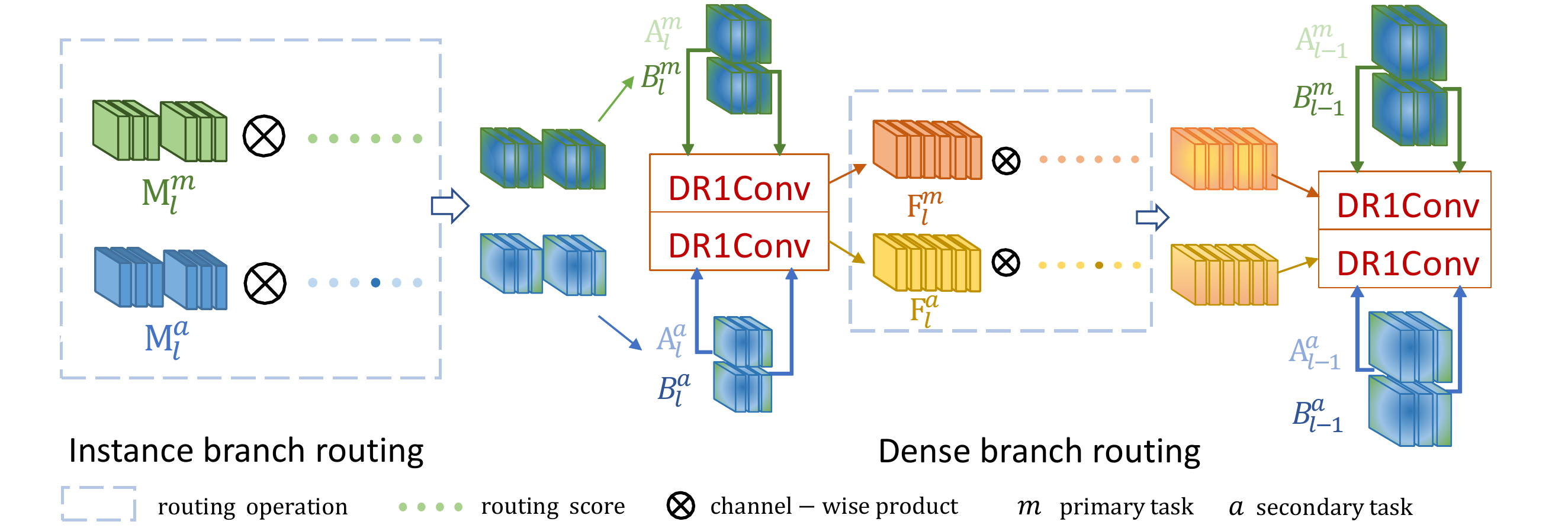}
\caption{The router generates routing scores for contextual information and dense feature map in instance and dense branches respectively. In the dense branch, the routed results are 2x upsampled and then used as the input of the next pyramid level. 
Best viewed in color.}
\label{fig:dr1basis}
\end{figure*}

Through the above three levels of information in our dynamic router, we could have a unified multi-task perception framework without sacrificing the performance of each task. The router only has one $1\times1$ convolution layer, leading to negligible computation cost. 
The overall computation graph for our proposed dynamic module is shown in Fig.~\ref{fig:dr1basis}.

\subsection{Task-specific prediction head}\label{sec:head}
In this work, we focus on three kinds of tasks, namely panoptic segmentation, monocular 3D object detection and depth estimation. We merge the instance-wise outputs ${\boldsymbol{e}^{(i)}}$ and dense features $ \mathbf F$ for both instance-level tasks (\eg, instance segmentation and 3D object detection) and dense prediction tasks (\eg, segmentation and depth estimation). The corresponding prediction heads are described as follows. 

The \textbf{panoptic segmentation} is the basic multitask system 
that we explore here. 
It jointly handles instance segmentation and semantic segmentation. 
Similar to other crop-then-segment models, we first crop a region of interest $\mathbf R^{(i)}\in\mathbb R^{D'\times56\times56}$ from the dense branch output $\mathbf F$ according to the detected bounding box $b^{(i)}$ using RoIAlign~\cite{he2017mask}. Then the crops are combined into the final instance-wise predictions guided by the instance embeddings ${\boldsymbol{e}^{(i)}}$. For segmentation and 3D detection, we split each $\boldsymbol{e}^{(i)} \in \mathbb R^{C^{seg}+C^{3
D}}$ into two vectors $\boldsymbol{e}^{(i)}_{seg} \in \mathbb R^{C^{seg}}$ and $\boldsymbol{e}^{(i)}_{3D} \in \mathbb R^{C^{3D}}$ respectively.

For panoptic segmentation, we use a new instance prediction module, called \textit{factored attention}, which has fewer parameters but can accept much wider basis features. We split the embedding into two parts 
$\boldsymbol{e}^{(i)}_{seg}= [\boldsymbol{t}^{(i)}:\boldsymbol{s}^{(i)}]$, where $\boldsymbol{t}^{(i)}$ is the projection kernel weights and $\boldsymbol{s}^{(i)}$ is the attention factors. First, we use $\boldsymbol{t}^{(i)}$ as the (flattened) weights of a $1\times 1$ convolution which projects the cropped bases 
$\mathbf R^{(i)}$ into a lower dimension tensor $\mathbf R'^{(i)}$ with width $K$:
\begin{equation}
    \mathbf R'^{(i)} = \boldsymbol{t}^{(i)} \ast \mathbf R^{(i)}
\end{equation}
where $\boldsymbol{t}^{(i)}$ is the reshaped convolution kernel with size $D'\times K$;
and 
$\ast$ is the convolution operator\footnote{This makes $\boldsymbol{t}^{(i)}$ a vector of length $D'K$.}. We choose $K=4$ to match the design choice of BlendMask~\cite{chen2020blendmask}.

We split 
$\boldsymbol{s}^{(i)}$ into $K$ diagonal matrix $\Sigma_k\in \mathbb R^{4\times4}$ and combine them with two learnable matrices $\mathbf U_k, \mathbf V_k\in \mathbb R^{4\times14}$ to generate $K$ attention  maps $\mathbf Q_k\in \mathbb R^{14\times14}$: 
\begin{equation}
    \mathbf Q^{(i)}_k = \mathbf U_k^\T\Sigma^{(i)}_k\mathbf V_k.
\end{equation}
Here, we 
set $\mathbf U_k$ and $\mathbf V_k$ as network parameters that are shared with all instances. This reduces the instance embedding parameters from $784$ to $16$ while still enabling us to form position-sensitive attention shapes.  $\mathbf R'^{(i)}$ and the full-attention $\mathbf Q^{(i)}$ are element-wise multiplied and summed along the first dimension to get the instance mask results. The outer product $\mathbf u_{kd}^T \mathbf v_{kd}$ of the $d$th row vectors in $\mathbf U_k$ and $\mathbf V_k$ can be considered as one of the components of $\mathbf Q_k$. We visualize all components learned by our network in Fig.~\ref{fig:uvbasis}.

We add minimal modifications to instance prediction for panoptic segmentation: a unified panoptic segmentation layer which is simply a $1\times1$ convolution $f_{pano}$ transforming the output $\mathbf F$ of the dense branch into panoptic logits with $C$ channels. The first $C_{stuff}$ channels are for semantic segmentation and the rest of $C_{thing}$ channels are for instance segmentation.

We split the weights for $f_{pano}$ along the columns into two matrix $\mathbf W_{pano} = [\mathbf W_{stuff}, \mathbf W_{thing}]$. The first $D'\times C_{stuff}$ parameters $\mathbf W_{stuff}$ are static parameters. $C_{stuff}$ is a constant equals to the number of stuff classes in the dataset, \eg, 53 for 
the 
COCO dataset.

The rest $D'\times C_{thing}$ parameters $\mathbf W_{thing}$ are dynamically generated. During training, $C_{thing}$ is the number of ground truth instances in the sample. For each instance $i$, there can be $N_i \ge 0$ predictions assigned to it with embeddings $\{\boldsymbol{e_n} \lvert n=1,\dots, N_i\}$ in the network assigned to it. 
For panoptic segmentation, we map them into a single embedding by computing their mean $\bar{\boldsymbol {e}}_i = \sum_n \boldsymbol{e}_n/N_i$. Then the $C_{thing}$ embeddings are concatenated into the dynamic weights $\mathbf W_{thing}$:
\begin{equation}
    \mathbf W_{thing} = [\bar{\boldsymbol {e}}_1, \bar{\boldsymbol {e}}_2, \dots, \bar{\boldsymbol {e}}_{C_{thing}}].
\end{equation}
The panoptic prediction can be computed with a matrix multiplication $\mathbf Y_{pano} = \mathbf W_{pano}^\T\mathbf F$. We combine thing and stuff supervisions and use cross-entropy loss for this prediction.

For \textbf{monocular 3D object detection}, we regress the 3D bounding box of each instance $i$ by predicting its 3D locations $loc = [c_x, c_y, z]$ encoded as 2.5D center offsets $c_x$ and $c_y$, corresponding depth $z$, dimensions $dim = [h, w, l]$  and its observation angle $\alpha$ encoded as $[sin\alpha, cos\alpha]$, for nuScenes dataset, an attribute label $a$ is also regressed. Accordingly, $\boldsymbol {e}^{(i)}_{3D}$ contains all 3D regression properties,
\begin{equation}
    \boldsymbol {e}^{(i)}_{3D} = [c_x, c_y, z_{inst}, h, w, l, \sin\alpha, \cos \alpha, a_1, \dots, a_A],
\end{equation}
where $a_1, \dots, a_A$ are the nuScenes attribute logits. To get the final instance depth prediction $z$, we add $z_{inst}$ to a densely predicted depth value from cropped bases $R$,
\begin{equation}
    z = z_{inst} + \operatorname{GAP}(R)^\top  w_z,
\end{equation}
where GAP is a global average pooling layer and $w_z \in \mathbb R^{D'}$
is a network parameter for dense depth prediction. We use the disentangled 3D corner regression loss similar to~\cite{liu2020smoke}:
\begin{equation}
     \mathcal{L}_{3D}= \mathcal{L}_{attr}(a_1, \dots, a_A) + \sum_{k \in {loc,dim,\alpha}}\lVert \hat{B_k}-B \rVert_1
\end{equation}
where $\hat{B_k}$ is the 3D bounding box coordinates predicted with $loc$, $dim$ and $\alpha$ respectively, $B$ is the ground truth box and $\mathcal{L}_{attr}$ is a classification cross entropy loss on nuScenes attributes.

For \textbf{monocular depth estimation}, we add three convolution layers to the output basis $\mathbf F$ to regress depth for every pixel, and $2\times$ interpolation operation after each convolution layer to upsample the output till to the original image size. 

Formally, the overall loss function of our multi-task framework can be formulated as,
\begin{equation}
\begin{split}
     \mathcal{L}= \sum{}(\mathcal{L}_{fcos}  & +  \lambda_{3D} \times (\mathcal{L}_{ctr}+\alpha  \mathcal{L}_{dim}+\mathcal{L}_{ori}+\beta \mathcal{L}_{loc})\\
    & + \mathcal{L}_{mask}+\mathcal{L}_{pano} + \mathcal{L}_{depth})
\end{split}
\end{equation}
where $\mathcal{L}_{fcos}$ is the original loss of FCOS, $ \mathcal{L}_{mask}$ and $ \mathcal{L}_{pano}$ denote the cross entropy loss that we used for panoptic masks. $\mathcal{L}_{depth}$ is L1 loss. $\lambda_{3D}$ = 0.4 which is set to balance the loss. $\alpha$ = 2, $\beta$ = 0.5, both empirically set through the single-task experiments.

\section{Experiments}

\subsection{Dataset and implementation details}

\smallskip
\noindent \textbf{Dataset} For pairwise and multi-task training, we select the Cityscapes dataset~\cite{cordts2016cityscapes, gahlert2020cityscapes}, which contains monocular 3D detection, depth, and panoptic segmentation annotations for 20 semantic categories related to urban scene understanding. The dataset comprises 5,000 finely annotated images, divided into 2,975 for training, 500 for validation, and 1,475 for testing.

In addition, we evaluate our basic two-branched framework on hybrid benchmark datasets. NuScenes~\cite{caesar2020nuscenes} contains 1.4M 3D object bounding boxes on 200K+ images over 83 logs. NuImages provides 700K segmentation masks on 93K images in more varied scenes (nearly 500 logs). 
These two datasets share the same 10 instance categories, and nuImages includes semantic masks for drivable surfaces.
we validate the efficacy of our approach in this partial label setting on nuScenes and nuImages datasets for joint segmentation and 3D detection. 

\smallskip
\noindent \textbf{Metric}
For panotic segmentation, we use  the standard panoptic quality (PQ) metric. For 3D object detection, the official detection score (DS) metric is used in Cityscapes \cite{gahlert2020cityscapes}. In nuScenes, the official evaluation metrics for the detection task are provided. The mean average precision (mAP) of nuScenes is calculated using the center distance on the ground plane rather than the 3D intersection over union (IoU) to align predicted results with ground truth. The nuScenes metrics also contain 5 types of true positive metrics (TP metrics), including ATE, ASE, AOE, AVE, and AAE for measuring translation, scale, orientation, velocity, and attribute errors, respectively. The nuScenes also defines a detection score (NDS) as $NDS= \frac{1}{10}[5mAP+ \sum_{mTP \in TP}({1 - \text{min}(1, mTP))]}$ to capture all aspects of detection tasks. 
The depth performance are measured by absolute relative error(Abs. Rel.), depth accuracy $\delta = max(\frac{d_{pred}}{d_{gt}}, \frac{d_{gt}}{d_{pred}})$ 
and root mean square error(RMSE).
\smallskip

\noindent \textbf{Data augmentation}\label{sec:aug}
Similar to SMOKE~\cite{liu2020smoke}, we regress a point that is defined as the projected 3D center of the object on the image plane. 
The projected keypoints allow us to fully recover the 3D location for each object with camera parameters. 
Let ${\left[ x \quad y \quad z \right]}^\T$~represents the 3D center of each object in the camera frame.
The projection of 3D points to points ${\left[ x_c \quad y_c \right] }^\T$~on the image plane can be obtained with the camera intrinsic matrix $K$ in a homogeneous form: 
\begin{equation}
    z_c \times{
    \begin{bmatrix}
    u \\
    v \\
    1 \\
    \end{bmatrix}=
    \begin{bmatrix}
    f_x &  ~~0  &  ~~u_0 \\
    0  & ~ ~ ~ f_y &~~ v_0 \\
    0  &~~  0  & ~~ 1 \\
    \end{bmatrix} \times
    \begin{bmatrix}
    x_c  \\
    y_c \\
    z_c \\
    \end{bmatrix}
    }
\end{equation}
When rescale and crop augmentation are used, it can be seen as the change in the camera intrinsic matrix. \eg, if we resize an image at a ratio of $s$, then crop it at $[x_0, y_0]$, the matrix is changed as follows:
\begin{equation}
    {
    \begin{bmatrix}
    f_x~~ & ~~u_0 \\
    f_y~~ & ~~v_0  \\
    \end{bmatrix}=
    \begin{bmatrix}
    f_x\times s ~~& ~~ u_0\times s - x_0 \\
    f_x\times s ~~& ~~ v_0\times s - y_0 \\
    \end{bmatrix}
    }
\end{equation}

To maintain the geometry consistency, we use relative crop ranging from [0.5, 1.0] of origin image size instead of the random crop. Using the above augmentations enables a stable training process.

\smallskip
\noindent \textbf{Implementation}
We implement our models based on the open-source project {\tt AdelaiDet}\footnote{\url{https://git.io/AdelaiDet}}. 
Unless 
specified otherwise, 
COCO pre-trained ResNet-50 is used as our backbone. We train our unified multi-task network on 8 Telsa V100 with batch size 32. The training schedule is 90k iterations and the learning rate is reduced by a factor of 10 at iteration 60K and 80K. Note that in single-task experiments, task-awareness information is disabled. For monocular depth estimation, we set the max regression distance to 120 meters.

\subsection{Main results}\label{sec:mainrel}

We compare our multi-task results with other state-of-the-art methods on Cityscapes benchmark. Table  \ref{table:cs-mtl} reports the results on panoptic segmentation, 3D object detection and depth estimation. 
Our method achieves significant improvements across all tasks. In particular, using a ResNet-50 backbone, D2BNet performs significantly better on 3D object detection and depth estimation tasks on Cityscapes test set, outperforming the previous best single-task methods by 3.8 DS and 0.09 depth accuracy. D2BNet surpasses the other multi-task method, MGNet, by 4.3 PQ and 1.37 RMSE on panoptic segmentation and depth estimation. 
We use official Panoptic-Deeplab open-source code and re-implement it by adding the same depth prediction module as ours, and an FCOS3D head to fit the multi-task setting. Panoptic-Deeplab suffers from task conflicts, while D2BNet outperforms it on three tasks, which is attributed to our dynamic modules.

\begin{table*}[htbp]
\centering
\small
\caption{Evaluation of multi-task prediction on Cityscapes. We use ResNet-50 as our backbone. The results of panoptic segmentation are on the validation set due to fewer methods reporting test set results.
The comparison results of 3D detection come from Cityscapes leaderboard.\protect\footnotemark[4] \dag~ uses Swin-S backbone \cite{liu2021Swin}. \dag~ uses ConvNeXt-B backbone\cite{liu2022convnet}.
}
\label{table:cs-mtl}
\resizebox{\textwidth}{!}{
\begin{tabular}{c|r|ccc|ccc|ccccc}
\hline
\multirow{2}{*}{}&  \multirow{2}{*}{Method}  & \multicolumn{3}{c|}{Panoptic Segmentation} & \multicolumn{3}{c|}{3D Detection} & \multicolumn{4}{c}{Depth Estimation}  \\
 &  & PQ & PQ\textsuperscript{Th} & PQ\textsuperscript{St} & DS & mAP & OS Yaw   &  Abs. Rel.$\downarrow$ & $\delta<1.25$  & $\delta<1.25^2 $ &$\delta<1.25^3$ & RMSE$\downarrow$\\
\hline
\hline
\multirow{9}{*}{single task}&Panoptic-DeepLab~\cite{cheng2020panoptic}  & 59.7 & - & - & - & - & - & - & - & - & - &-\\
& UPSNet~\cite{xiong2019upsnet}  & 59.3  & 54.6  & 62.7 & - & - & - & - & - & - & - &-  \\
& SOGNet~\cite{yang2020sognet} & 60.0  & 56.7 &  62.5 & - & - & - & - & - & - & - &-  \\
& 3D-GCK~\cite{gahlert2020single} & - & - & - & 37.4&	42.5&	81.9 & - &-&-&-&-\\
& AVPNet2.3  & - & - & - & 40.1&	43.5&	88.0  & - &-&-&-&-\\
& iFlytek & - & - & - & 42.9 & 47.6 &	80.4  & - &-&-&-&-\\
& Xu~\cite{Xu_2018_Padnet} &-&- & - &-&-&-&0.246& 0.786& 0.905& 0.945&7.117\\
& SDC-Depth~\cite{wang2020sdc}  &-&-& - &-&-&-&0.227& 0.801& 0.913& 0.950 & 6.917 \\
& Saeedan~\cite{Saeedan_2021_WACV} &-&-& - &-&-&-&0.178& 0.771& 0.922& 0.971 &-\\
\hline
\multirow{5}{*}{multi-task}
&Panoptic-DeepLab$\ast$  & 56.6 & 54.1 & 59.3 & 43.8 & 48.6 & 85.0 & 0.143 & 0.792 & 0.930 & 0.977 &7.052\\
& MGNet~\cite{schon2021mgnet} & 55.7  & - &  - & - & - & - & - & - & - & -  & 8.3 \\
& PanopticDepth\cite{gao2022panopticdepth} & 57.0  & 52.3&  60.5 & - & - & - & - & - & - & -  &\textbf{ 6.69}\\
& PanopticDepth\dag~\cite{gao2022panopticdepth} & 60.4  & 56.0 &  63.6  &  -  &  -  &  -  &  -  &  -  &  - &  -  & -\\
& D2BNet &  60.0 & 57.2 & 62.0  &  46.7 &50.0 &90.4 & 0.095 & 0.897 &0.956& 0.991  & 6.931 \\
&  D2BNet\ddag &   \textbf{61.8} &  \textbf{59.4} &  \textbf{63.9}  &    \textbf{48.3} & \textbf{52.5} & \textbf{92.3}  & \textbf{ 0.082} & \textbf{0.912} & \textbf{0.973}&  \textbf{0.996 } &  6.70 \\
\hline
\end{tabular}}
\end{table*}
\footnotetext[4]{\url{https://www.cityscapes-dataset.com/benchmarks/\#3dbbox-results}}

\begin{figure}[t!]
\centering
\includegraphics[width=0.5\textwidth]{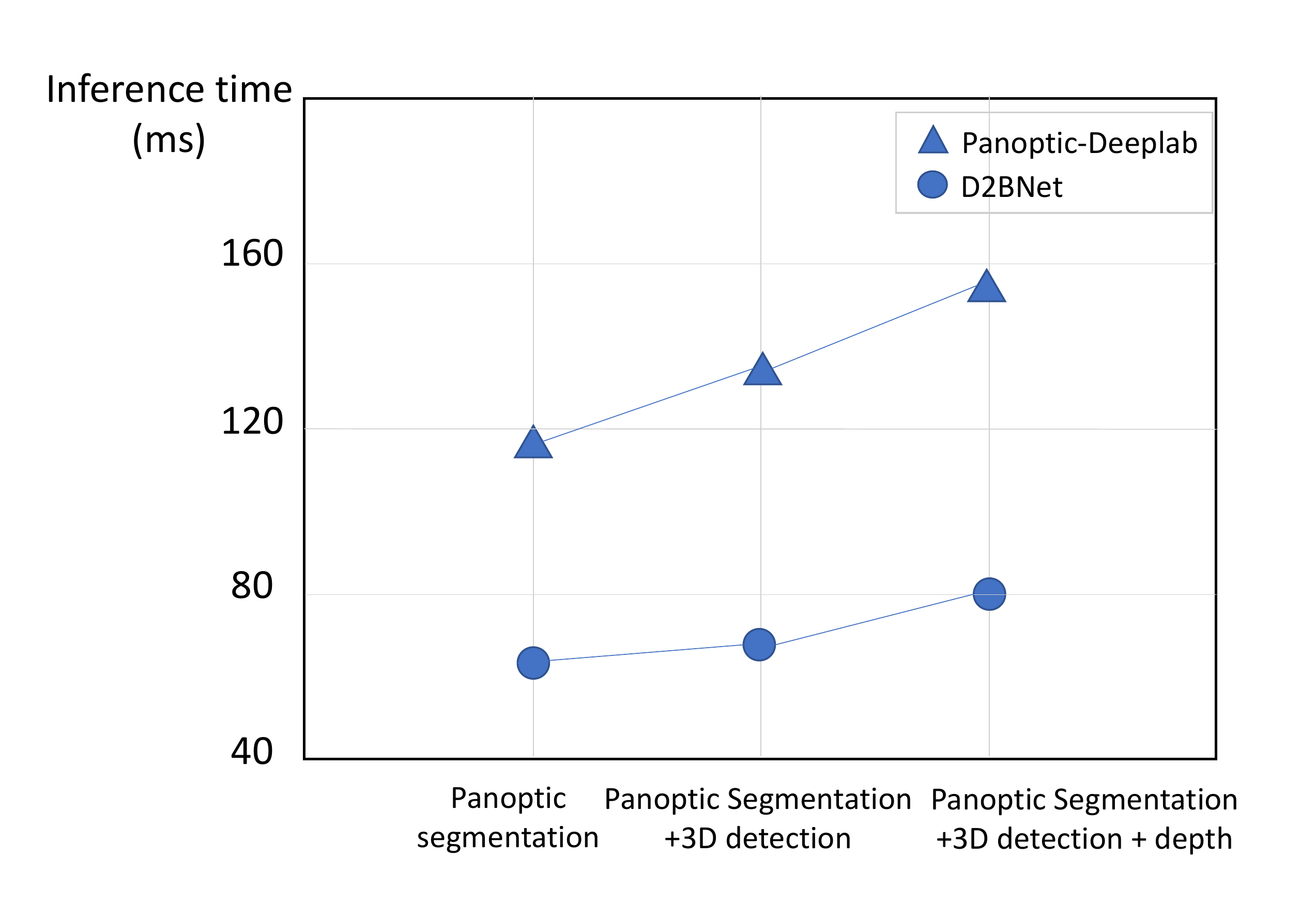}
\caption{Comparision of inference time(ms) for different multitask frameworks on Cityscapes. 
}
\label{fig:inf_time}
\end{figure}

In addition, We compare the inference time of D2BNet with the multi-head framework Panoptic-DeepLab for a single panoptic segmentation task, panoptic segmentation with 3D detection task, and joint three tasks. We add a separate FCOS3D~\cite{wang2021fcos3d} head to Panoptic-DeepLab models for 3D detection. The inference time is measured with the ResNet-50 backbone in batch size 1. In Panoptic-DeepLab, the final stage is dilated. The input resolution is 1024 $\times$ 2048. 
Computation statistics for different frameworks are shown in Fig. \ref{fig:inf_time}. Our model saves significant computation time by reusing most features across multiple tasks.

\subsection{Ablation experiments}
In this section, we conduct an empirical study of the design choices for the three levels of awareness in our dynamic module and the entire two-branched architecture under a multi-task setting using different benchmark datasets.

\smallskip
\noindent \textbf{Dynamic modules in multi-task network}  We evaluate the effectiveness of our dynamic modules under the multi-task, full label setting by adding them to the baseline. The experiment is implemented on Cityscapes dateset. As shown in Table  \ref{table:dy-cs}, each module is beneficial for every task, and combining all three levels of awareness achieves the best results on all tasks at once.

\begin{table}[t]
\caption{Ablation studies of dynamic modules in our two-branched network on Cityscapes validation set.}
    \centering
    \small
    \label{table:dy-cs}
    \begin{tabular}{ccc|ccc}
    \hline
    DMP & TDR & CDR &  DS  &  PQ   &Abs. Rel. \\  
    \hline
    \hline
    $\times$ & $\times$   & $\times$ &  45.38   & 56.82    &  0.146  \\
    \checkmark & $\times$   & $\times$  &  45.73 & 57.01   &  0.140  \\
    \checkmark   & \checkmark & $\times$ & 47.09  & 58.13  &  0.117   \\
    \checkmark   & $\times$   & \checkmark & 47.6  & 57.19  &  0.122   \\
     \checkmark & \checkmark &  \checkmark & \textbf{48.66} & \textbf{58.6}  & \textbf{0.117}   \\
    \hline
      \end{tabular}
\end{table}

\noindent \textbf{Pairwise joint training on Cityscapes} To have an intuitive understanding of the mutual influence of co-training tasks, we disable our task- and channel-aware modules, then train three tasks pairwise, the results are shown in Table  \ref{table:pairwise}. To preserve geometric prior, depth and 3D detection tasks are trained using relative crop, therefore panoptic segmentation is trained in identical augmentation, which sacrifices nearly 1.8 points on panoptic quality metric compared to random crop. In our experiments, both 3D detection and depth estimation tasks benefited from co-training with segmentation, while the segmentation task co-trained with depth resulted in noticeable performance degradation.

\begin{table}[htbp]
\centering
\small
\caption{Single task training and pairwise co-training results without our task- and channel-aware dynamic modules on Cityscapes validation set. Each column is the same task metric and the corresponding row is the task it co-trained with. The diagonal line in the upper three rows represents the single-task training results for reference. The last row displays the results of training with our three dynamic modules in a multi-task setting. 
Note that segmentation results are trained under the relative crop augmentation instead of random crop, which sacrifices nearly 1.8 points on panoptic quality metric.
}
\label{table:pairwise}
\resizebox{0.5\textwidth}{!}{
\begin{tabular}{cc |ccc}
\hline
 & & 3D (DS) & Segmentation(PQ)  & Depth(Abs. Rel.)\\
\hline
\hline
\multirow{3}{*}{\rotatebox{90}{T With}} &3D    &49.79 & 57.70 & 0.122 \\
 &Segm  & 50.62 & 58.23 & 0.111 \\
& Depth & 50.9  & 56.40 & 0.124 \\
\hline
& D2BNet & 50.87  & 58.40 & 0.114 \\
\hline
\end{tabular}}
\end{table}

\noindent \textbf{3D object detection and segmentation from partial labels on nuScenes} To demonstrate the efficacy of our multi-task framework on other benchmarks, we train it on nuScenes and nuImages datasets. Since images in nuScenes and nuImages are not overlapping, we are facing a missing label problem for joint training.  we experiment with three different settings: (i) single-task training, (ii) alternate training where two datasets are joint and each batch contains data from a single task, (iii) pseudo-labeling where we train on nuScenes 3D detection dataset with segmentation annotations generated by a single-task model trained on nuImages. We apply different augmentations for different tasks, if the batch has 3D detection annotations, we only apply horizontal flip, otherwise, we apply flip and random resize with short-size from $[720, 1080]$. Results are shown in Table  \ref{table:3dsegm}, both joint training methods have positive a impact on the 3D detection task. For alternate training, we resample nuScenes and nuImages with ratio 1: 2, for individual task, the training iterations is about halved, which could probably explain the performance drop in mask AP compared to single-task training. 

\begin{table}[htb]
\centering
\small
\caption{Multi-task training over 3D object detection and instance segmentation. 3D mAP is the nuScenes official distance-based mAP.
}
\label{table:3dsegm}
\begin{tabular}{r |ccc}
\hline
Method & Dataset & 3D mAP  & mask mAP \\
\hline
\hline
single-task & nuScenes & 28.79& -\\
single-task & nuImages  &-  &42.28\\
alternate training & NS+NI & 30.63 & 39.81\\
pseudo-labeling & nuScenes & \textbf{30.82} & -\\
\hline
\end{tabular}
\end{table}

\noindent \textbf{The performance of 3D object detection with segmentation labels}\label{sec:3d-results} 
We compare our model to the previous best vision-only methods for 3D object detection on nuScenes. Although these two datasets are not explicitly linked, nuImages could contain similar scenes in nuScenes datasets. To avoid including external data, we only use the COCO pretrained model on nuImages to generate panoptic segmentation pseudo labels on nuScenes. We use a ResNet-101 backbone with deformable convolutions on the last two stages with interval 3 and train for 450k iterations with batch size 16. Quantitative results are shown in Table  \ref{table:3dsegmall}. Our solution achieves 1st place on the fourth nuScenes 3D detection challenge in the vision-only track.

\begin{table*}[t]
\centering
\small
\caption{Multi-task training over 3D object detection and instance segmentation. Comparing to
FCOS3D, our backbone has 2/3 less DCNs. TTA is test-time augmentation, which we keep identical to FCOS3D.
}
\label{table:3dsegmall}
\resizebox{0.8\textwidth}{!}{
\begin{tabular}{r |cccccccc}
\hline
Method & Dataset &  TTA  & mAP$\uparrow$  & mATE$\downarrow$  & mASE$\downarrow$ &  AOE$\downarrow$ &  mAAE$\downarrow$  & NDS$\uparrow$ \\
\hline
\hline
CenterNet~\cite{zhou2019objects}(DLA) &  val &   & 30.6 &  0.716  & 0.264 &  0.609  & 0.658  & 0.328\\
FCOS3D~\cite{wang2021fcos3d}  &  val  & \checkmark &  34.3  & 0.725 &  0.263  & 0.422 &  0.153  & 0.415\\
 PersDet~\cite{zhou2022persdet}  &   val  &  &   34.6  &  0.660 &   0.279  &  0.540 &   0.207  &  0.408\\
D2BNet       & val  &  & \textbf{35.4}  & 0.729  & 0.271  & 0.363 &  0.175 &  \textbf{0.423}\\

\hline
Noah CV Lab  &  test &  &  33.1 &  0.660  & 0.262 &  0.354  & 0.198  & 0.418\\
FCOS3D       &  test &  \checkmark  & 35.8  & 0.690  & 0.249 &  0.452 &  0.124 &  0.428\\
D2BNet        & test  & \checkmark  & \textbf{36.3} &  0.667  & 0.259 &  0.402 &  0.120 &  \textbf{0.437}\\
\hline
\end{tabular}}
\end{table*}

\subsection{Qualitative results on Cityscapes}\label{sec:visresult}
\begin{figure*}[htb]
\centering
\includegraphics[width=0.98\textwidth]{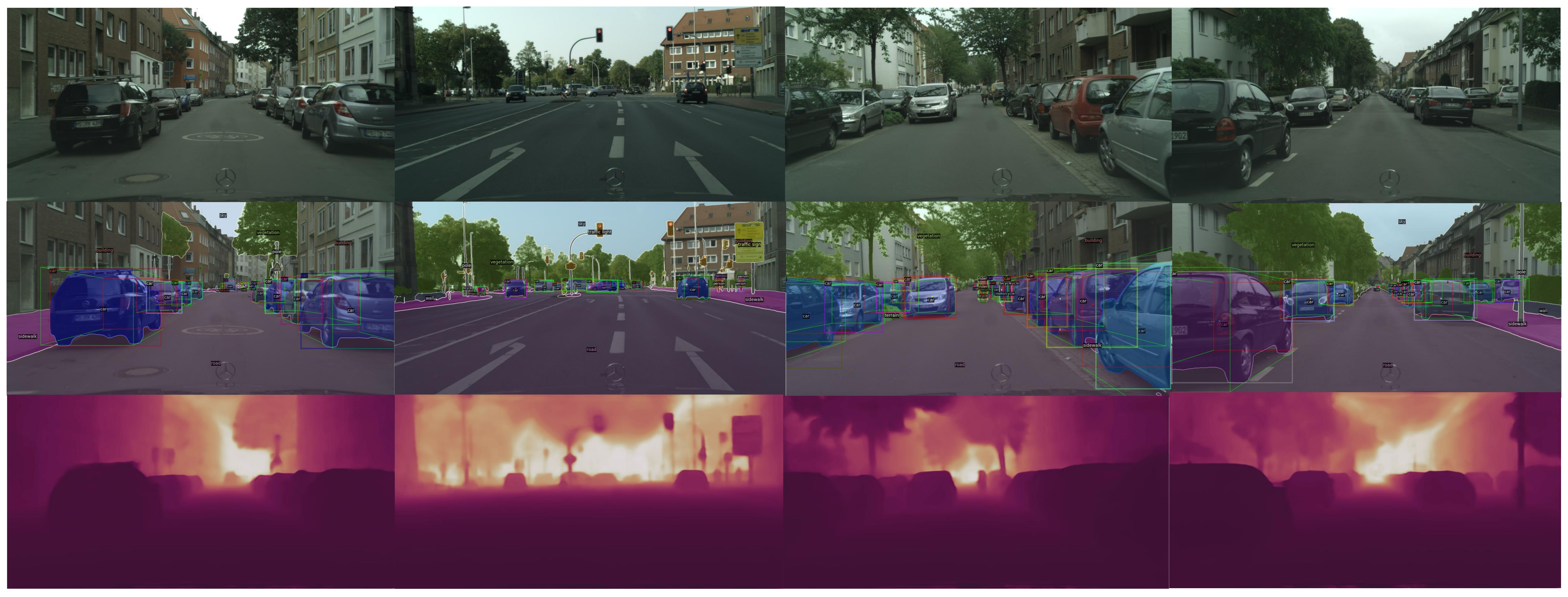}
\caption{Qualitative analysis of our multi-task results on Cityscapes validation set. The first row is raw images and the last two rows are our multi-task predictions. Best viewed in color and zoom in.
}
\label{fig:result}
\end{figure*}

We demonstrate some qualitative results in Fig. \ref{fig:result} on Cityscapes dataset. For a clear visualization, multi-task predictions are shown in the last two rows. The second row is our predictions on 2D object detection, 3D object detection and panoptic segmentation, and the visualized results of depth estimation are shown in the last row.

\subsection{Relations between panoptic segmentation and depth estimation}
\begin{figure*}[htbp]
\centering
\includegraphics[width=0.8\textwidth]{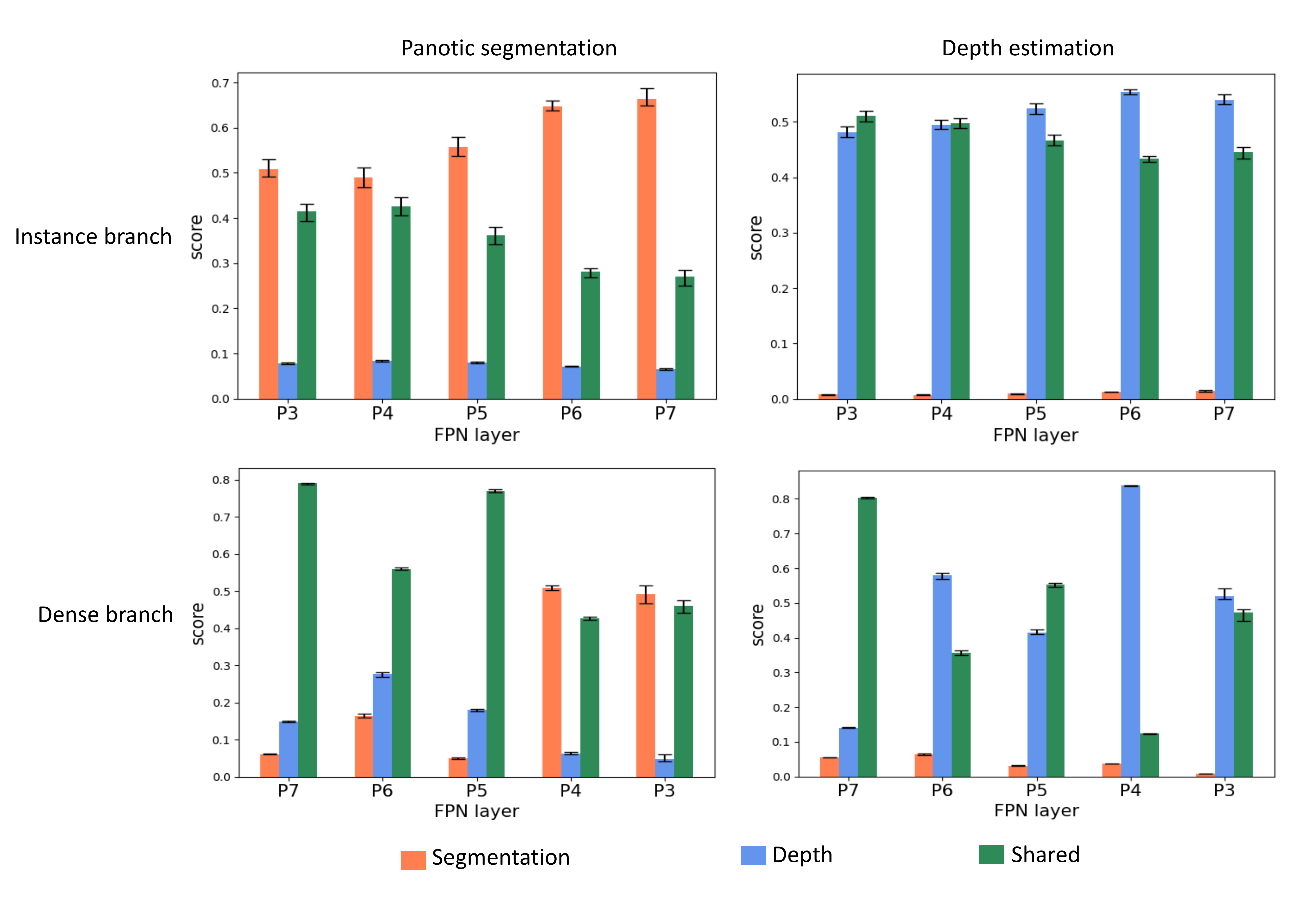}
\caption{The quantitative relationship of the feature sharing between panoptic segmentation and depth estimation across Cityscapes validation set. In addition to two task branches, we introduce a shared branch to enable sharing relationships. Each subfigure displays the routing score on the vertical axis and different layers of FPN on the horizontal axis. The upper sub-figures show routing scores on the instance branch for representing segmentation and depth, respectively. The bottom two subfigures show segmentation and depth scores on the dense branch.
}
\label{fig:tskgate}
\end{figure*}

In the early stage of devising our sharing network, we attempt to figure out the relationship among dense prediction tasks. Simply combining tasks in a unified network results in performance degradation. To investigate the sharing relationship, we designed experiments to co-train panoptic segmentation and depth estimation in separated branches with additional shared parameters. The weighting scores on $\{\mathbf M_l\}$ in the instance branch and $\{\mathbf F_l\}$ in DR1Conv projections in the dense-branch are visualized in Fig. \ref{fig:tskgate}, 
Our co-training on panoptic segmentation and depth estimation tasks revealed information sharing on both the instance and dense branches, with varying degree of feature sharing across branches and feature scales. 
Taking panoptic segmentation and depth estimation as instance, tasks share contextual features in the instance branch and the sharing extents have no significant changes across different FPN layers, while it shows a decreased tendency of sharing as the size of the dense feature map increases. As a result, we set routers both on different FPN levels and branches.

\subsection{Results of our D2BNet on single tasks}
In this section, we take the routers away and validate the effectiveness of our proposed model for single-task on different benchmarks.

\noindent \textbf{The performance of panoptic segmentation}\label{sec:segm} 
We compare D2BNet with recent panoptic segmentation networks on the COCO \code{test}-\code{dev} split. We increase the training iterations to 270K ($3 \times$ schedule), tuning the learning rate down at 180K and 240K iterations. 
The running time is measured on the same machine with the same setting. We use multi-scale training with shorter side randomly sampled from $[640, 800]$. We run the models with batch size 1 on the whole COCO val2017 split using one GTX 1080Ti GPU. We calculate the time from the start of model inference to the time of final predictions, including the post-processing stage. Results on panoptic segmentation are shown in Table  \ref{table:panoptic}. Our model achieves the best speed-accuracy trade-off and is two times faster than the mainstream separate frameworks. Particularly, the running time bottleneck for UPSNet~\cite{xiong2019upsnet} is the stuff/thing prediction branches and the final fusion stage, which makes the R-50 model almost as costly as the R-101 DCN model. Our method is faster than Panoptic FCN because our instance prediction module is more efficient. 

\begin{table*}[t!]
\centering
\small
\caption{\textbf{Panoptic results} on COCO. R-50 models are evaluated on {\tt  val2017 split} and R-101 models are evaluated on test-dev. All models are evaluated with the official code and the best models publicly available on the same machine. Panoptic-DeepLab does not provide trained models on COCO. We measure its speed by running the Cityscapes pretrained model on COCO val2017. Models with * have deformable convolutions in the backbone. All hyperparameters are set to be the same with BlendMask~\cite{chen2020blendmask}. 
}
\label{table:panoptic}
\resizebox{0.75\textwidth}{!}{
\begin{tabular}{r |c|c|ccc|cc}
\hline
Method &Backbone & Time (ms) & PQ & SQ & RQ & PQ\textsuperscript{Th} & PQ\textsuperscript{St} \\
\hline
\hline
Panoptic-FPN~\cite{kirillov2019panoptic} &\multirow{6}{*}{R-50} & 89 & 41.5 & 79.1 & 50.5 & 48.3 & 31.2 \\
UPSNet~\cite{xiong2019upsnet} & & 233 & 42.5 & 78.2 & 52.4 & 48.6 & 33.4 \\
SOGNet~\cite{yang2020sognet} & & 248 & 43.7 & 78.7 & 53.5 & 50.6 & 33.1 \\
Panoptic-DeepLab~\cite{cheng2020panoptic} & & 149 & 35 & - & - & - & - \\
BlendMask~\cite{chen2020blendmask} & & 96 & 42.5 & 80.1 & 51.6 & 49.5 & 32.0 \\
D2BNet & & \textbf{79} & 42.9 & 79.8 & 52.0 & 49.5 & 32.9 \\
\hline
Panoptic-DeepLab & Xception-71 & - & 41.4 & - & - & 45.1 & 35.9\\
\hline
Panoptic-FPN \cite{kirillov2019panoptic} &\multirow{5}{*}{R-101} & 111 & 43.6 & 79.7 & 52.9 & 51.0 & 32.6\\
BlendMask & & 117 & 44.5 & 80.7 & 53.8 & 52.1 & 33.0\\
UPSNet$^*$ & & 237 & \textbf{46.3} & 79.8 & \textbf{56.5} & 52.7 & 36.8 \\
\textbf{D2BNet} & & 99 & 44.5 & 80.7 & 53.8 & 51.7 & 33.5\\
\textbf{D2BNet}$^*$ & & 109 & 46.1 & \textbf{81.5} & 55.3 & \textbf{53.1} & 35.5 \\
\hline
\end{tabular} }
\end{table*}

We also compare the \textbf{instance segmentation} results on nuImages dataset. On nuImages, we train with the 1$\times$ schedule and random resize with short size [720, 1080] and $512 \times 1024$ crop augmentations. There are no panoptic evaluation protocols for nuImages, so we compare our model with Mask R-CNN implemented in MMDet{\tt}\footnote{\url{https://github.com/open-mmlab/mmdetection3d/tree/master/configs/nuimages}}. Models both use pretrained weights on COCO.

\begin{table}[htb]
\small
\centering
\begin{tabular}{r|ccc}
\hline
Method & AP & AP$_{50}$ & AP$_{75}$ \\
\hline
\hline
Mask R-CNN & 40.5 &   \textbf{70.6} &    40.7  \\
D2BNet &  \textbf{42.3}   &   68.3   &  \textbf{44.0} \\
\hline
\end{tabular}
\caption{Instance segmentation results on the nuImages validation set. Mask R-CNN result is from mmdetection3d official implementation. Both models use ResNet-50 backbone pre-trained on COCO and are trained with the 1$\times$ schedule.}
\label{table:nuimages}
\end{table}

\begin{table*}[htb]
\centering
\small
\resizebox{0.75\textwidth}{!}{
\begin{tabular}{r |c|c|ccc|ccc}
\hline
Method &Backbone & Time (ms) & AP &AP$_{50}$ &AP$_{75}$ &AP$_S$ &AP$_M$ &AP$_L$\\
\hline
\hline
Mask R-CNN~\cite{he2017mask}&\multirow{4}{*}{R-50} &74 & 37.5 & 59.3 & 40.2 & 21.1 & 39.6 & 48.3 \\
BlendMask~\cite{chen2020blendmask}& & 73 & 38.1 & 59.5 & 41.0 & 21.3 & 40.5 & 49.3 \\
CondInst~\cite{tian2020conditional} & &  72 & 38.7 & 60.3 & 41.5 & 20.7 & 41.0 & 51.3\\
D2BNet & & \textbf{69} & 38.3 & 59.6 & 41.2 & 21.1 & 40.4 & 50.0 \\
\hline
Mask R-CNN \cite{he2017mask}&\multirow{5}{*}{R-101} & 94 & 38.8 & 60.9 & 41.9 & 21.8 & 41.4 & 50.5 \\
BlendMask& & 94 & 39.6 & 61.6 & 42.6 & 22.4 & 42.2 & 51.4\\
CondInst & & 93 & 40.1 & 61.9 & 43.0 & 21.7 & 42.8 & 53.1\\
\textbf{D2BNet} & & 89 & 39.8 & 61.6 & 42.9 & 21.9 & 42.4 & 51.9\\
\textbf{D2BNet}$^*$ & & 98 & \textbf{41.2} & \textbf{63.2} & \textbf{44.5} & \textbf{22.6} & \textbf{43.8} & \textbf{54.7} \\
\hline
\end{tabular}}
\caption{\textbf{Instance segmentation results} on COCO \code{test}-\code{dev}. Models with * have deformable convolutions in the backbone.}
\label{table:main}
\end{table*}

We compare D2BNet with recent \textbf{instance segmentation} networks on the COCO \code{test}-\code{dev} split. We increase the training iterations to 270K ($3 \times$ schedule), tuning the learning rate down at 180K and 240K iterations. All instance segmentation models are implemented with the same code base, \code{Detectron2}\footnote{\url{https://github.com/facebookresearch/detectron2}} and 
The running time is measured on the same machine with the same setting. We use multi-scale training with shorter side randomly sampled from $[640, 800]$. Results on 
instance segmentation are shown in Table  \ref{table:main}.

\subsection{Additional ablation results on single tasks}\label{sec:addresult}

\noindent \textbf{Effectiveness of dynamic factors in location-aware module}  DR1Conv has two dynamic components $\mathbf A$ and $\mathbf B$. And each of them has the effect of channel-wise modulation pre-/post- convolution respectively. By removing both of them, our basis module becomes a vanilla FPN. We train networks with each of these two components masked out. The second row in Table  \ref{table:dr1conv-pano} shows that DR1Conv can improve both the thing and stuff segmentation qualities. The combination of these two dynamic factors yields higher improvement than the increments of the two factors individually added together.

\begin{table}[t]
\caption{Panoptic segmentation results on Cityscapes with the dynamic factors removed.
}
\centering
\small
\label{table:dr1conv-pano}
\begin{tabular}{r |ccc}
\hline
Method & PQ & PQ\textsuperscript{Th} & PQ\textsuperscript{St}\\
\hline
\hline
w/o DMP & 58.7   &  55.9  &  60.2  \\
w/ DMP & \textbf{60.0} & \textbf{57.2} &  \textbf{62.0}   \\
\hline
\end{tabular}
\end{table}
\textbf{Context feature position} The contextual information $\mathbf M$ is computed with the features from the box tower of the FCOS~\cite{tian2019fcos}, a crucial difference from self-attention and squeeze-and-excite blocks. To examine this effect, we move the top layer for contextual information computation to the FPN outputs and class towers, which both badly hurts the segmentation performance,  AP$_{75}$ especially, even worse than the vanilla baseline without dynamism. Results are shown in the third row and Table  \ref{table:position}. This proves that the correspondence between instance embedding and contextual information is important. 

\textbf{Channel width of the dense branch} Choosing a proper channel width of the dense branch is also important for panoptic segmentation accuracy. A more compact basis output of size 32 does not affect the class agnostic instance segmentation result but will lead to much worse semantic segmentation quality, which has to discriminate 53 different classes. To accurately measure the influence of different channel widths and make sure all models are fully trained, we train different models with the 3x schedule.  Doubling the channel width from 32 to 64 can improve the semantic segmentation quality by 2.1. Results are shown in Table  \ref{table:width}.

\textbf{Border padding} We also notice that border padding can affect the performance of semantic segmentation performance. The structure difference between our dense branch and common semantic segmentation branch is that we have incorporated high-level feature maps with strides 64 and 128 for contextual information embedding. We assume that this leads to a dilemma over the padding size. A smaller padding size will make the features spatially misaligned across levels. However, an overly large padding size will make it very inefficient. Making an $800\times800$ image divisible by $128$ will increase $25\%$ unnecessary computation cost on the borders. We tackle this problem by introducing a new upsampling strategy with is spatially aligned with the downsampling mechanism of strided convolution and reduce the padding size to the output stride, \textit{i.e.}, 4 in our implementation. Results are shown in Table  \ref{table:padding}, our aligned upsampling strategy requires minimal padding size while being significantly better in semantic segmentation quality PQ\textsuperscript{St}.

\begin{table}[t]
\small
\centering
\begin{tabular}{r|ccc}
\hline
Position & AP & AP$_{50}$ & AP$_{75}$ \\
\hline
\hline
None & 34.7   &  55.5      &  36.8    \\
FPN & 34.2   &  55.5      &  36.0    \\
class tower & 34.5 & 55.6 &  36.5  \\
box tower & \textbf{35.2}   &   \textbf{56.1}     &  \textbf{37.5}   \\
\hline
\end{tabular}
\caption{Instance segmentation results with the contextual information from different positions.}
\label{table:position}
\end{table}

\begin{table}[t]
\small
\centering
\begin{tabular}{r|c|ccc}
\hline
Attention & Time (ms) & AP & AP$_{50}$ & AP$_{75}$ \\
\hline
\hline
Vector & \textbf{68.7} & 35.2   & 56.1 &  37.5    \\
Full &72.0 & 36.2 &  56.7  & 38.7  \\
Factored & 69.2 & \textbf{36.3}   &   \textbf{56.9}     &  \textbf{38.8}   \\
\hline
\end{tabular}
\caption{Comparison of different instance prediction modules. Vector is channel-wise vector attention in YOLACT~\cite{bolya2019yolact}; full is the 3D full attention tensor in BlendMask~\cite{chen2020blendmask} and factored is the factored attention introduced in Sec.  \ref{sec:head}.}
\label{table:attention}
\end{table}

\begin{figure}[htbp]
\centering
\includegraphics[width=0.45\textwidth]{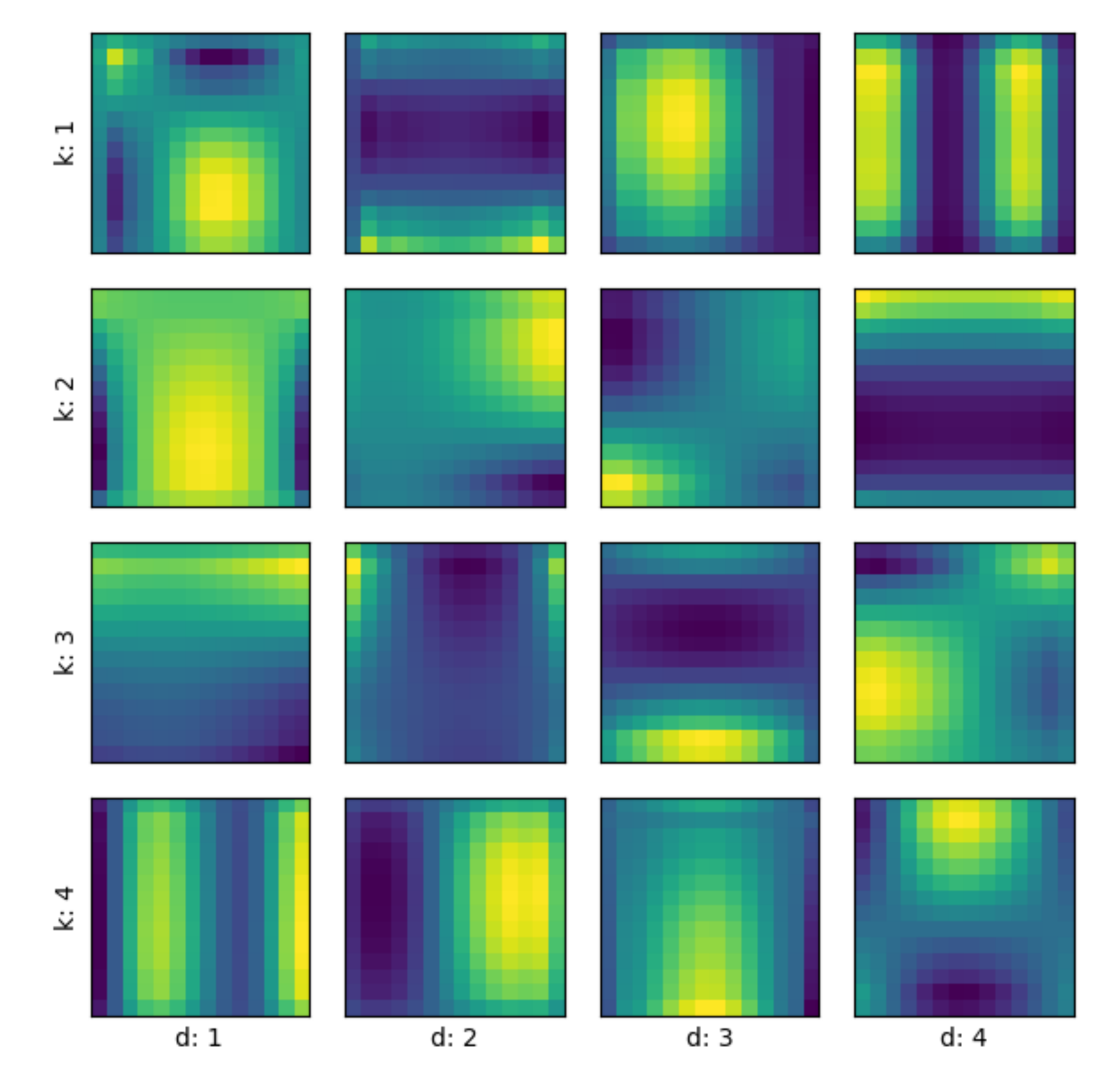}
\caption{\textbf{Attention components} $k$ is the index of the bases and $d$ is the index of the attention factors. The attention map at $(k, d)$ is the $k$th attention map in $\mathbf Q$ generated by a one-hot instance embedding $\mathbf s$ with the $d$th element valued $1$.
}
\label{fig:uvbasis}
\end{figure}

\textbf{Efficiency of the factored attention} We compare the performance and efficiency of different instance prediction modules 
in Table  \ref{table:attention}. Our factored attention module is almost as efficient as the channel-wise modulation and can achieve the best performance. In addition, we visualize all components learned by our network in Fig \ref{fig:uvbasis}.

\textbf{Position sensitive attention for panoptic segmentation} Unfortunately, even though beneficial for instance segmentation, we discover that position sensitive attention has a negative effect on panoptic segmentation. It enforces the bases to perform position-sensitive encoding on all classes, even for stuff regions, which is unnecessary and misleading. The panoptic performance for different instance prediction modules are shown in the fourth row at Table  \ref{table:prediction}. Using factored attention makes the semantic segmentation quality drop 
by 
2.6 points.  Thus, in our panoptic segmentation and multi-task experiments, the factor attention is only used in thing regions, the stuff classes are directed using convolutional layers to predict masks.
We also study the dense module channel width choices and border padding effect on the segmentation performance. As a result, we choose channel width 64 for panoptic segmentation and aligned upsampling to reduce the border padding size.

\begin{table}[t]
\centering
\small
\begin{tabular}{r |ccc}
\hline
Width & PQ & PQ\textsuperscript{Th} & PQ\textsuperscript{St}\\
\hline
\hline
32 & 41.8 &  49.1  &  30.7  \\
64 & \textbf{42.9} & \textbf{49.5} &  \textbf{32.9}   \\
128 & 42.8 & 49.5 & 32.8 \\
\hline
\end{tabular}
\caption{Comparison of different channel widths in the dense branch for panoptic segmentation. All models are with a ResNet-50 backbone and are trained with the 3x schedule. 
}
\label{table:width}
\end{table}

\begin{table}[t]
\centering
\small
\begin{tabular}{r |ccc}
\hline
Divisibility & PQ & PQ\textsuperscript{Th} & PQ\textsuperscript{St}\\
\hline
\hline
32 & 39.5 &  46.5  &  28.8  \\
128 & 39.9 & 46.5 & \textbf{30.0}  \\
4 w/ aligned & \textbf{40.0} & \textbf{46.8} &  29.9  \\
\hline
\end{tabular}
\caption{Comparison of different padding strategies for panoptic segmentation. The baseline method is padding to 32x, divisibility of C5 from ResNet. Padding to 128x is for the divisibility of the dense branch. 4 w/ aligned is padding the input size to 4x and applying our aligned upsampling strategy.
}
\label{table:padding}
\end{table}

\begin{table}[t]
\centering
\small
\begin{tabular}{r |ccc}
\hline
Method & PQ & PQ\textsuperscript{Th} & PQ\textsuperscript{St}\\
\hline
\hline
Vector & \textbf{40.0} & \textbf{46.8} &  \textbf{29.9} \\
Factored & 39.0 & 46.8 & 27.3  \\
\hline
\end{tabular}
\caption{Position sensitive attention for panoptic segmentation. Vector is the baseline model with vector instance embeddings. 
}
\label{table:prediction}
\end{table}

\section{Conclusion}

In this work, we propose a Dynamic Two-Branched Network (D2BNet) for multi-task perception, targeting to share features as much as possible and leverage the common representation among tasks. We break down tasks into two branches, using instance and dense branches to extract higher- and lower-level information, respectively, We then apply task-specific prediction heads for the final predictions. Cross-branch information communication is performed with a lightweight dynamic operation, DR1Conv. Meanwhile, use a task- and channel-wise dynamic router to isolate task-specific features and utilize the common properties of tasks. The benefits are twofold: a structure with better feature-sharing properties lays the foundation for joint instance-wise and dense prediction multi-task learning research, while also reducing the computation cost in real-world applications.

\section*{Data Availability Statement}

Datasets 
used in this work 
are all publicly available. NuScenes and NuImages datasets are available at {\tt https://www.nuscenes.org/}. Cityscapes is available at 
{\tt https://www.cityscapes-dataset.com/}, and MSCOCO at
{\tt https://cocodataset.org/}.

\section*{Acknowledgement}
This work was supported by National Key R\&D Program of China (No.\ 2020AAA0106900), the National Natural Science Foundation of China (No.\ U19B2037, No.\ 62206244), Shaanxi Provincial Key R\&D Program (No.\ 2021KWZ-03), Natural Science Basic Research Program of Shaanxi (No.\ 2021JCW-03).

\bibliographystyle{IEEEtran}
\bibliography{reference}

\end{sloppypar}
\end{document}